\pdfoutput=1

\documentclass[11pt]{article}

\usepackage[]{acl}

\usepackage{times}
\usepackage{latexsym}

\usepackage{graphicx}

\usepackage[T1]{fontenc}

\usepackage[utf8]{inputenc}

\usepackage{microtype}

\usepackage{calligra}
\usepackage{algorithm}
\usepackage{algpseudocode}
\usepackage{amsmath}
\usepackage{marvosym}
\title{MoDS: Model-oriented Data Selection for Instruction Tuning}

 \author{Qianlong Du\textsuperscript{1}, Chengqing Zong\textsuperscript{1,2} \and Jiajun Zhang\textsuperscript{1,2,3,\Letter}\\
         \textsuperscript{1}Institute of Automation, Chinese Academy of Sciences \\
         \textsuperscript{2}School of Artificial Intelligence, University of Chinese Academy of Sciences\\
           \textsuperscript{3}Wuhan AI Research\\
         \texttt{\{qianlong.du,cqzong,jjzhang\}@nlpr.ia.ac.cn} \\}

\begin{document}
\maketitle
\begin{abstract}
Instruction tuning has become the de facto method to equip large language models (LLMs) with the ability of following user instructions. Usually, hundreds of thousands or millions of instruction-following pairs are employed to fine-tune the foundation LLMs. Recently, some studies show that a small number of high-quality instruction data is enough. However, how to select appropriate instruction data for a given LLM is still an open problem.
To address this problem, in this paper we present a model-oriented data selection (MoDS) approach, which selects instruction data based on a new criteria considering three aspects: quality, coverage and necessity. 
First, our approach utilizes a quality evaluation model to filter out the high-quality subset from the original instruction dataset, and then designs an algorithm to further select from the high-quality subset a seed instruction dataset with good coverage. The seed dataset is applied to fine-tune the foundation LLM to obtain an initial instruction-following LLM. Finally, we develop a necessity evaluation model to find out the instruction data which are performed badly in the initial instruction-following LLM and consider them necessary instructions to further improve the LLMs. In this way, we can get a small high-quality, broad-coverage and high-necessity subset from the original instruction datasets. Experimental results show that, the model fine-tuned with 4,000 instruction pairs selected by our approach could perform better than the model fine-tuned with the full original dataset which includes 214k instruction data. Codes, data, and models are available\footnote{https://github.com/CASIA-LM/MoDS}.

\end{abstract}

\section{Introduction}

With the development of artificial intelligence, large language models, such as GPT-3 \cite{brown2020language}, GPT-4 \cite{GPT4}, PaLM \cite{chowdhery2022palm}, OPT \cite{zhang2022opt}, and some other open-source LLMs \cite{llama,vicuna2023,touvron2023llama}, have showed revolutionary potential in general language understanding and generation. As a critical technique of LLMs, instruction tuning \cite{ouyang2022training,shu2023exploitability,wang2023aligning,wang2022self,cao2023instruction,li2023self,wang2023openchat} enables LLMs to correctly follow various kinds of user instructions.

In early researches, instruction tuning \cite{wang2022self,xu2023wizardlm,yu2023large,sun2023principle,ding2023enhancing} mainly focuses on how to construct large-scale, diverse, and high-quality instruction data. Recently, \cite{zhou2023lima} proposes a LIMA model which demonstrates that only 1,000 carefully crafted high-quality instructions can enable the model to possess a powerful instruction-following capability. Their results suggest that almost all of the knowledge in LLMs has been learned during pre-training, and only a small number of instruction tuning data is required to activate models to follow instructions and produce high quality responses. Subsequently, there has been a growing interest among researchers in the systematic filtration of high-quality and comprehensive subset from the extensive pool of instruction dataset \cite{cao2023instruction,chen2023alpagasus,li2023quantity}. 
However, these data filtration methods rely too much on extra LLMs or mainly focus on the quality of instructions. Different from those methods, this paper proposes a model-oriented approach which selects instruction data based on a new criteria considering three aspects: quality, coverage and the necessity as well. The quality requires the selected instruction data to be good enough for both questions and answers. The coverage requires the selected instruction data to be diverse enough. The necessity indicates that the selected instruction data indeed fill the ability gap for the LLM of interested.

In order to select high-quality instruction data from a large dataset, this paper first proposes to use a quality evaluation model to assess all the (instruction, input, output) triplets, and then filter out the instruction data with high-quality scores. After that, we further propose to use a k-center greedy algorithm \cite{sener2017active} to select instruction data from the high-quality subset. This k-center greedy algorithm could select a subset of data points that are the farthest apart, thereby making the instruction data we collect are diverse and have broader coverage. In this way, we can get a seed instruction dataset for the target LLM fine-tuning. Due to the difference of pre-training data, model architecture and training processes, different LLMs vary in their abilities, which result in the fact that different LLMs require different kinds of instruction data. In order to further find out the instruction data the specific LLM needed, we fine-tune the given LLM with the seed instruction dataset, and then assess its inference results on all the high-quality instruction dataset. In this way, we can filter out the instructions on which the specific LLM performs poorly, making up an augmented dataset for the target LLM. This augmented dataset indicates the instruction-following capabilities that LLM lacks. Finally, by merging the seed instruction data and the augmented data, we get a high-quality, broad-coverage and high-necessity subset from the original large-scale instruction datasets. We then utilize these selected data to fine-tune the target LLM again. 

Our contributions can be summarized as follows:

(1) We propose a new criteria for instruction data section including quality, coverage and necessity, and verify that they are valuable for the LLM fine-tuning.

(2) We propose a model-oriented instruction selection approach which not only considers the quality and coverage of instruction data, but also integrates the necessity of instructions based on the ability of specific LLMs.

(3) Experimental results show that the LLM fine-tuned with 4,000 instruction data selected by our approach could achieve a better performance than the LLM fine-tuned with the full original dataset (214k), indicating that our approach is effective in selecting valuable instruction data with high-quality, broad-coverage and high-necessity.

\section{Related Work}

Recent researches show that instruction tuning could enable LLMs to be tailored to specific domains, tasks, or applications by providing explicit instructions or guidelines \cite{wei2021finetuned,cao2023instruction}. In order to enhance the instruction-following abilities of LLMs, previous work mainly focus on increasing the data sizes through various strategies \cite{honovich2022unnatural,wang2022self,alpaca,kopf2023openassistant,honovich2022unnatural}. However, the work of \citeauthor{zhou2023lima} (\citeyear{zhou2023lima}) illustrates that even a small number of constructed high-quality instructions could empower the model with a powerful instruction-following capability. They indicate that most of the knowledge in LLMs have been acquired during the pre-training procedure, and only a limited number of instruction data are enough to activate LLMs to follow instructions and generate high-quality responses. Their work demonstrates significant improvements compared to LLMs which are fine-tuned with similar-scale unfiltered data. However, it should be noted that their approach requires manual involvement to select data from extensive datasets, which is both time-consuming and costly.

Motivated by the work of \cite{zhou2023lima}, \citeauthor{cao2023instruction} (\citeyear{cao2023instruction}) proposed an instruction mining approach which adopts a linear quality rule and bag of indicators to evaluate the quality of instruction-following data. However, they do not conduct comparisons with LLMs trained on the complete dataset, and their approach is very complex.

Besides, \citeauthor{chen2023alpagasus} (\citeyear{chen2023alpagasus}) recently propose a ALPAGASUS model which directly leverages an external LLM (chatgpt) to score each instruction and then selects 9k Alpaca data with a threshold. Their model surpasses the performance of the official Alpaca model which is trained on the complete dataset. However, they rely excessively on external LLMs with great performance.

Different from them, \citeauthor{li2023quantity} (\citeyear{li2023quantity}) present a self-guided approach for LLMs to independently identify and choose relevant instruction pairs from extensive open-source datasets. In their approach, they introduce an Instruction-Following Difficulty (IFD) metric as a tool to identify gaps in a model's responses versus its autonomous generation capability. It signficantly reduces the need for manual curation and the associated costs for instruction tuning. However, when computing the IFD metric they only adopt one answer for each instruction, which neglects that the responses for each instruciton are diverse. Besides, they don't pay much attention to the quality and coverage of instruction data during selection procedure.

\begin{figure*}[!t]
	\centering
	\includegraphics[width=0.98\linewidth]{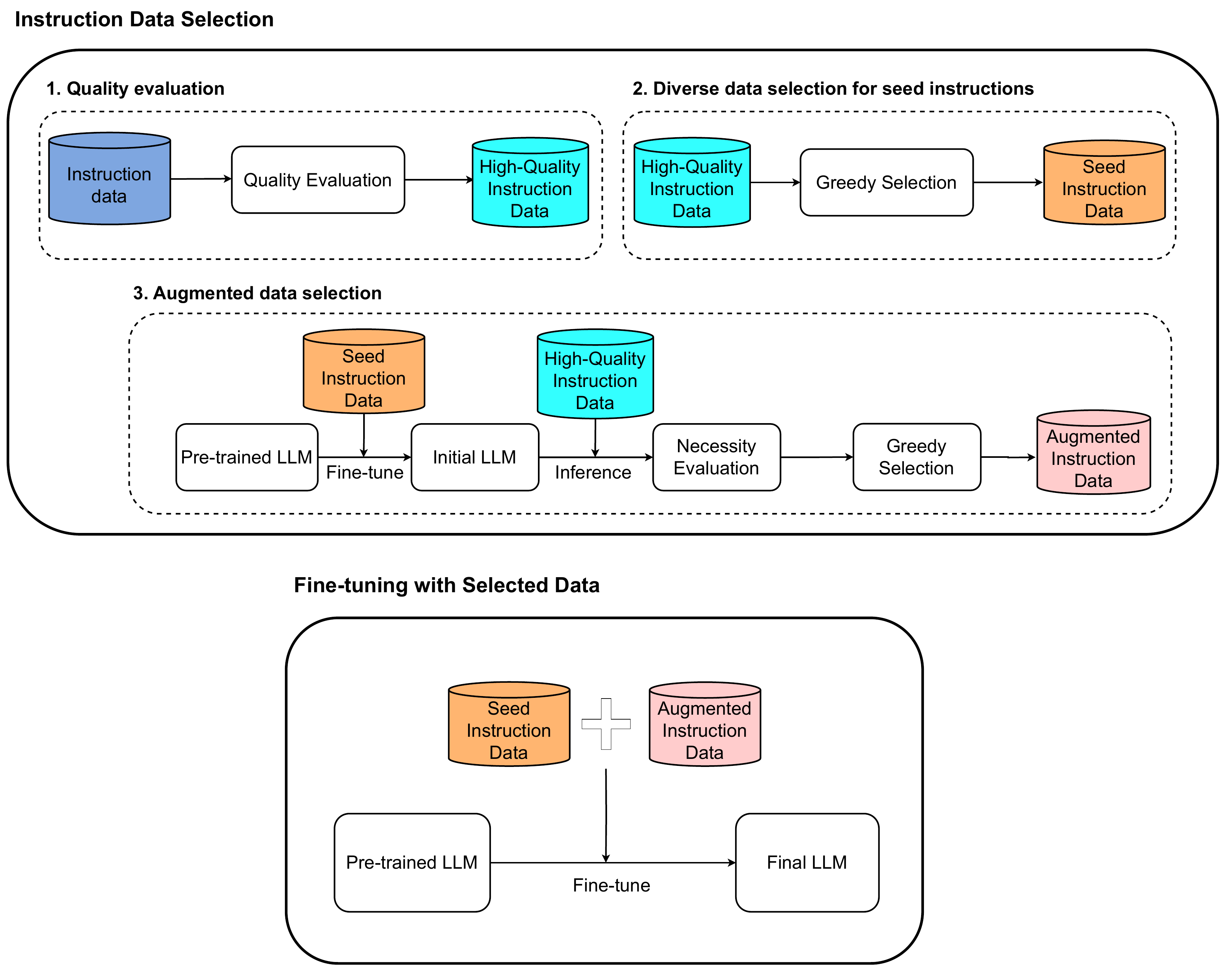}
	\caption{The architecture of MoDS approach. The top part shows the procedure of instruction data selection, while the bottom part illuminates the fine-tuning procedure.}
	\label{fig:architecture}
\end{figure*}

\section{Methodology}

\subsection{Which instructions are the valuable data for a given LLM}

\citeauthor{zhou2023lima} (\citeyear{zhou2023lima}) show that LLM's knowledge has been mostly learnt during pre-training. Instruction tuning is mainly used to teach a given LLM to learn how to follow a certain pattern when interacting with human, and only a small number of carefully crafted high-quality instructions are enough to equip the given LLM with powerful instruction-following capabilities. However, for different LLMs, as the knowledge and abilities they have learnt during the pre-training procedure are different, the instruction tuning data they require shoud be different as well. Consequently, how to select the most crucial data for a given LLM has garnered much attention of researchers. After analyzing some LLMs and instructions, we find that the valuable instruction tuning data for one given LLM are mainly decided by the following three aspects:

\textbf{Quality}. "Quality" refers to the data quality of both the instructions and their corresponding responses in the dataset, which directly influences the knowledge LLM learns. As demonstrated in the work of \cite{zhou2023lima,chen2023alpagasus,cao2023instruction}, high-quality instruction data can effectively enhance LLM's ability to follow instructions.

\textbf{Coverage}. 
"Coverage" refers to the types of instrucitons the dataset includes. It represents the diversity of one instruction dataset. The more diverse instruction the dataset covers, the greater the potential of stimulating the capabilities of a large language model is. Researches of \cite{iyer2023optiml,wang2023far,wang2022supernaturalinstructions,longpre2023flan} also show that enhancing the diversity of instruction data can effectively enhance LLM's ability to follow instructions during fine-tuning.

\textbf{Necessity}. "Necessity" indicates the importance and uniqueness of one instruction for fine-tuning a specific LLM. As described in the work of \cite{zhou2023lima}, LLMs have already acquired a substantial amount of knowledge and capabilities during pre-training. Instruction tuning primarily focuses on how to use a limited number of instruction data to stimulate LLM's capabilities, enabling LLMs to follow a certain pattern when interacting with human. Due to the knowledge and capabilities LLMs have learned are different, the importance and uniqueness of the same instruction data may vary for different LLMs.
For a given instruction, if the LLM could generate a high-quality response, it indicates that the LLM has already owned the ability of following this type of instructions, and this instruction data is non-essential for the fine-tuning. Conversely, if the LLM cannot generate a good response for that instruction, it suggests that the LLM lacks the ability to follow this type of instructions, and that instruction is necessary for optimizing the LLM's capabilities.

\subsection{Instruction Data Selection}

As mentioned in the previous section, the process of selecting effective instruction data from a large-scale dataset for a given LLM is primarily determined by three aspects: {\bf{quality}}, {\bf{coverage}}, and {\bf{necessity}}. To efficiently select the most valuable instruction data with these three aspects, this paper proposes a model-oritented approach for instruction data selection, which is shown in the top of Figure \ref{fig:architecture}. This approach mainly includes three modules: {\bf{Quality Evaluation}}, {\bf{Diverse Data Selection for Seed Instructions}} and {\bf{Augmented Data Selection}}. The details are presented in the following.

\begin{figure*}[!t]
	\centering
	\includegraphics[width=\linewidth]{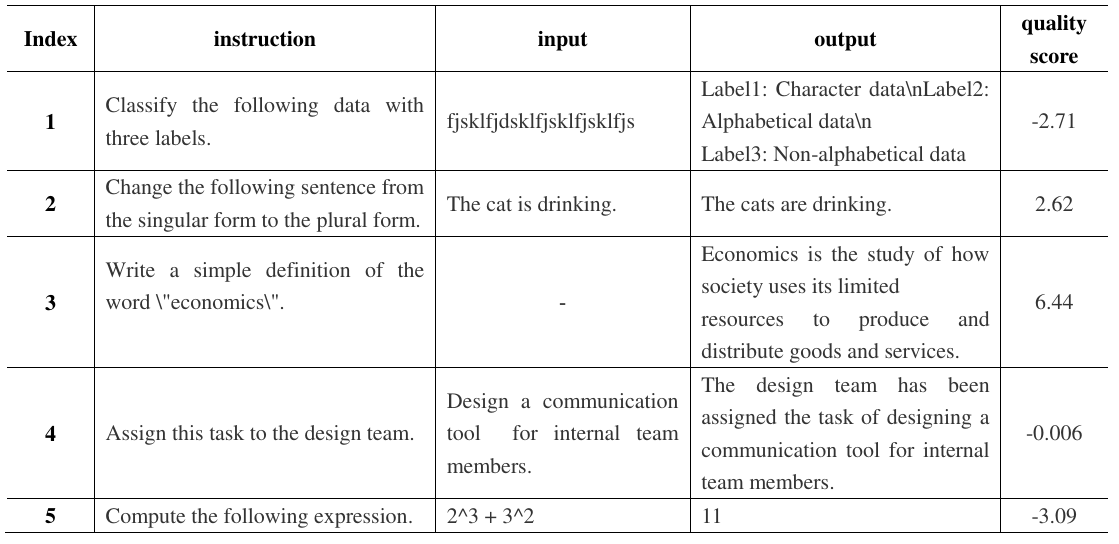}
	\caption{Examples of instruciton data with quality scores.}
	\label{fig:quality-score}
\end{figure*}

\subsubsection{Quality Evaluation}

The quality of instruction data plays a crucial role in the learning of instruction-following capabilities for LLMs. Therefore, to select effective instruction data, we first evaluate the qualities of instruction data and their corresponding response in the large-scale dataset, and then filter out the higher-quality data from it. When assessing the qualities of instruction data, we utilize the reward-model-deberta-v3-large-v2\footnote{https://huggingface.co/OpenAssistant/reward-model-deberta-v3-large-v2} model which is developed by OpenAssistant. This is a reward model designed based on the DeBERTa \cite{he2021deberta} architecture, and is trained on four different types of human feedback data \cite{nakano2022webgpt,stiennon2022learning,bai2022training}, endowing it with the abilities of QA model evaluation, reward scoring, and detecting potential toxic response via ranking. In this paper, we mainly adopt its reward scoring capability to generate a quality score for each (instruction, input, output) triplet in the large-scale dataset. As shown in Figure \ref{fig:quality-score}, some examples with quality scores are displayed.

 After generating the quality scores for each (instruction, input, output) triplet, we will filter them with a threshold $\alpha$. Through collecting the (instruction, input, output) triplet whose quality score is larger than $\alpha$, we can get a high-quality instruction dataset.

\subsubsection{Diverse Data Selection for Seed Instrucitons}

 \begin{algorithm}
  \caption{K-Center-Greedy \cite{sener2017active}}
  \textbf{Input:} data $x_i$, existing pool $s^0$ and a budget $b$
  Initialize $s=s^0$
  
  \textbf{repeat}
  
  $u = argmax_{i\in [n]\textbackslash s}min_{j\in s}\Delta (x_i,x_j) $
  
  $s = s \cup {u}$
  
  \textbf{Until} $|s| = b + |s^0|$
 
  \textbf{return} $s \textbackslash s^0$
 \end{algorithm}

After getting a high-quality instruction dataset, we will further select data from it. In order to select diverse instruction data with the maximum coverage, we propose to use K-Center greedy algorithm \cite{sener2017active} for data selection. K-Center greedy algorithm is proposed by \cite{sener2017active} in 2017, which is a simple yet effective approach used to address the K-Center problem. The objective of the K-Center problem is to choose a subset of K centers from a given set of data points in a manner that minimizes the maximum distance between any data point and its nearest center. This algorithm commences by selecting an initial center, typically the point farthest from any existing centers, and then proceeds to add new centers iteratitively. At each step, it chooses the point farthest from the current set of centers. As shown in Algorithm 1 \cite{sener2017active}, it presents the details of this algorithm.

During diverse data selection process, we generate the sentence embeddings for all instructions with BERT \cite{devlin2018bert}, which are used to compute the distances of different data points. Through this module, we can get a seed instruction dataset which has a great diversity and broad coverage.

\subsubsection{Augmented Data Selection}

For different LLMs, as the knowledge and capabilities they learned in the pre-training procedure are different, the instruction tuning data they require will be different as well. For one instruction, if the given LLM could generate a good response, it indicates that the given LLM has owned the ability to handle this type of instruction, and this instruction data is not necessary for the fine-tuning of the LLM. Conversely, if the LLM cannot generate a good response, it suggests that the LLM couldn't effectively process that type of instruction data, and the instruction data is very important and unique for the fine-tuning of the target LLM.

In section 3.2.2, we have generated a seed instruction dataset with high-quality and broad-coverage. However, as the valuable instructions vary for different LLMs, the seed instruction dataset may not include all the instructions the target LLM needs. In order to find out these missed instructions, we first fine-tune the pre-trained LLM with the seed instruction dataset, generating an initial LLM. Then we generate the responses of all the instructions in high-quality dataset with the initial LLM. After that, we will use a necessity evaluation model to compute a review score for each instruction and its generated response. In this paper, we still adopt the reward model used in section 3.2.1 as the necessity evaluation model. If the review scores are less than the threshold $\beta$, it represents that the initial LLM could not generate good responses for these instructions, and it doesn't own the capabilities to handle that types of instructions. After collecting all the instructions with low review scores, we will again use the K-center greedy selection algorithm described section 3.2.2 to select a subset from them, and then build an augmented dataset. This dataset could effectively compensate for the capability deficiencies of the initial LLM.

\subsubsection{Fine-tuning with Selected Instruction Data}

Following the methods outlined in the previous section, we can get a seed instruction dataset and its augmented dataset for a given LLM. After that, we will merge these two datasets, and then fine-tune the raw pre-trained LLM. This process has been shown in the bottom part of Figure \ref{fig:architecture}. In this way, we can get the final LLM which has a good instruction-following capability. The raw pre-trained LLM used in this paper is LLaMA 2 \cite{touvron2023llama}.

\section{Experiments}

\subsection{Datasets}
\subsubsection{Training set}

\textbf{Alpaca}. In this paper, we use Alpaca \cite{alpaca} which is built by Stanford University as one of the original instruction datasets. This dataset comprises 52,002 (instruction, input, output) triplets. It was created using the self-instruct approach \cite{wang2022self} with ChatGPT. And the LLM trained on this dataset shows a good instruction-following ability. However, relying too much on ChatGPT make researches concern about the quality of instruction data.

\noindent\textbf{Mixture Dataset}. In addition to Alpaca, we also build a much larger mixture instruction dataset as the original training data. In this dataset, we mix the instruction data from HC3 \cite{guo-etal-2023-hc3}, alpaca \cite{alpaca}, alpaca-evol-instruct \cite{xu2023wizardlm}, dolly-v2 \cite{DatabricksBlog2023DollyV2}, InstructWild \cite{instructionwild} and lima \cite{zhou2023lima},  and then construct a mixture instruction dataset which includes 214,526 (instruction, input, output) triplets. Compared to Alpaca, this dataset contains more diverse and rich instructions ranging from open-domain, medical, legal, financial and so on.

\subsubsection{Test set}

In order to evaluate the performance of our proposed approach, we also utilize five different test sets as in the work of \cite{li2023quantity}, including Koala \cite{vu2023koala}, WizardLM \cite{xu2023wizardlm}, Self-instruct \cite{wang2022self}, Vicuna \cite{vicuna2023} and LIMA \cite{zhou2023lima}. 
These test sets contain 180, 218, 252, 80 and 300 human-curated instruction data respectively, covering math, coding, writing, knowledge, computer and other domains.

\subsection{Detatils of Training and Testing}

\textbf{Training details.} 
In this paper, we adopt LLaMA 2 \cite{touvron2023llama} with 7B parameters as the raw LLM for fine-tuning. During fine-tuning procedure, we utilize the same hyperparameters as the work of \cite{alpaca}, which include a learning rate of 2e-5, a warmup ratio of 0.03, a weight decay of 0.0 and a batch size of 128. Besides, the fine-tuning epoch is set to 3. And we conduct all fine-tuning and evaluation experiments on NVIDIA RTX A100. During the procedure of quality evaluation and necessity evaluation, both of the threshold $\alpha$ and $\beta$ is set to 0.0 for Alpaca dataset, while they are set to 1.0 and -1.0 respectively for Mixture dataset.  

\noindent\textbf{Testing details.} 
During testing process, human evaluation is the most accurate and reliable approach to evaluate the instruciton-following capabilities of LLMs. However, this approach is very time-consuming and costly. Moreover, the evaluation results may also be effected by human biases. Consequently, in this paper, we also utilize ChatGPT and GPT-4 for the evaluation of LLMs as in the work of \cite{zhou2023lima,chen2023alpagasus,li2023quantity}.
During evaluation process, all the LLMs are prompted to generate the responses for all of the instructions in test sets. Subsequently, the evaluation LLM is prompted to assign a score for each of these responses based on the aspects of relevance and accuracy. And the score is on a scale from 1 to 10. Besides, in order to eliminate the impact of positional bias on the judements, following the work of \cite{chen2023alpagasus,li2023quantity}, we also evaluate the responses of two given LLMs twice, but with different ordering in the prompts. Finally, we will compare their scores in these two times of evaluations respectively, and the criteria of winning is presented in the following:

\textbf{Wins}: If the model outperforms in both comparions or wins in one while tying in the other.

\textbf{Tie}: If the model ties in both comparions or wins in one while losing in the other.

\textbf{Loses}: If the model loses in both comparisons or ties in one while losing in the other.

\subsection{Results and Analysis}

\begin{figure}[!t]
	\centering
	\includegraphics[width=\linewidth]{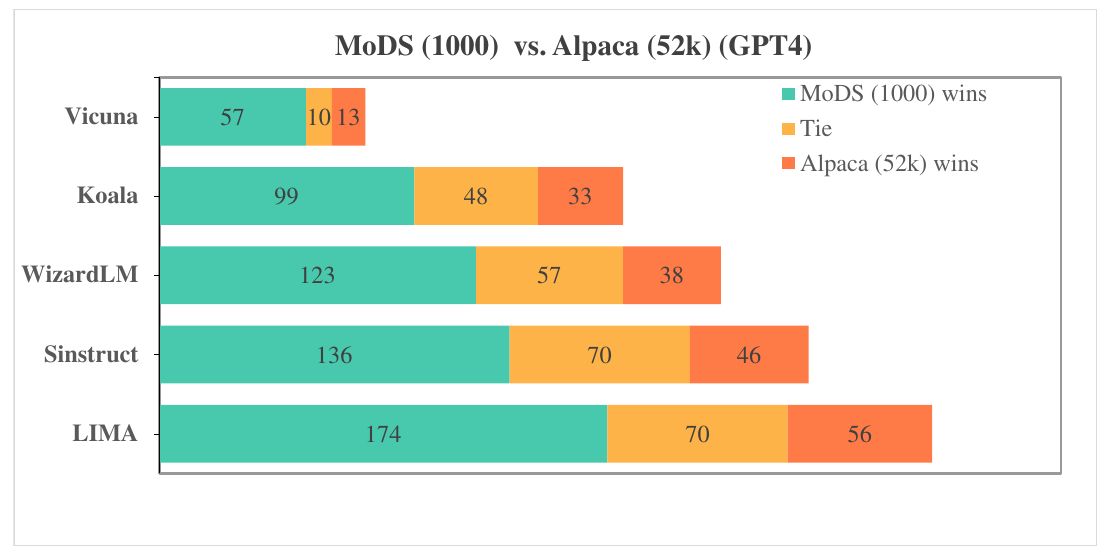}
	\caption{The comparison of our MoDS model trained on selected data with the model trained on full Alpaca dataset.}
	\label{fig:mods-vs-self}
\end{figure}

\begin{figure*}[!t]
	\centering
	\includegraphics[width=\linewidth]{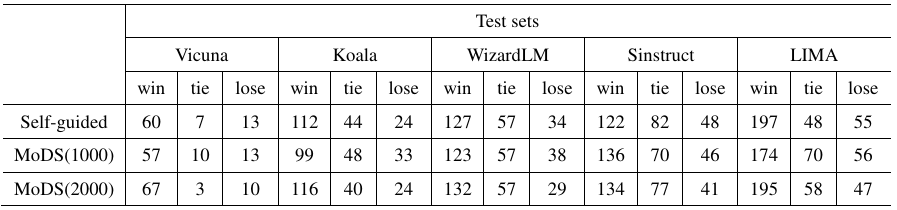}
	\captionof{table}{The comparisons between different instruction selection models and full Alpaca model on different test sets, using GPT-4 as the judge.}
	\label{fig:mods-vs-self-guided}
\end{figure*}

This section mainly present the performance of our approach on different test sets. As shown in Figure \ref{fig:mods-vs-self}, it presents the comparison of our MoDS model with the model trained on the full alpaca dataset. During fine-tuning procedure, both the size of seed instruction dataset and augmented instruction dataset of our MoDS model are 500. From this figure we can see that our MoDS approach which only adopts 1000 instruction data achieves a better performance than the model trained on the full alpaca dataset, which utilizes 5,2000 instructions. This results indicate that our instruction data selection approach is effective, and a small number of high-quality, broad-coverage and high-necessity selected instruction data could also make LLMs have a powerful instruction-following ability.

In order to compare our method with the self-guided instruction data selection approach proposed by \cite{li2023quantity}, Table \ref{fig:mods-vs-self-guided} shows their comparisons with the corresponding models trained on full Alpaca dataset. In the work of \cite{li2023quantity}, they introduce a Instruction-Following Difficulty (IFD) metric as a tool to identify gaps in a model's responses versus its autonomous generation capability and then select 5\% percentage (about 2600 instructions) of the full alpaca data to fine-tune the raw LLM. In Table \ref{fig:mods-vs-self-guided}, Self-guided represents the model fine-tuned with the 2600 instruction data\footnote{https://github.com/MingLiiii/Cherry\_LLM} selected by self-guided approach. And MoDS(1000) represents the model fine-tuned with 500 seed instruction data and 500 augmented instruction data which are choosen by our approach, while MoDS(2000) represents the model fine-tuned with 1000 seed instruction data and 1000 augmented instruction data. For all of them, the pre-trained language model is LLaMA 2. From this table, we can see that MoDS(1000) is comparable to Self-guided on Vicuna, Koala, WizardLM and LIMA test sets, while it is better than Self-guided on Sinstruct test set. And MoDS(2000) is better than Self-guided on all of the test sets. It should be noted that the numbers of instruction data utilized by MoDS(1000) and MoDS(2000) are smaller than the Self-guided model. The results demonstrate that our model-oriented approach can better select instruction data the target LLM needs, and then effectively enhance LLM's instruction-following capabilities.

\begin{figure}[!t]
	\centering
	\includegraphics[width=\linewidth]{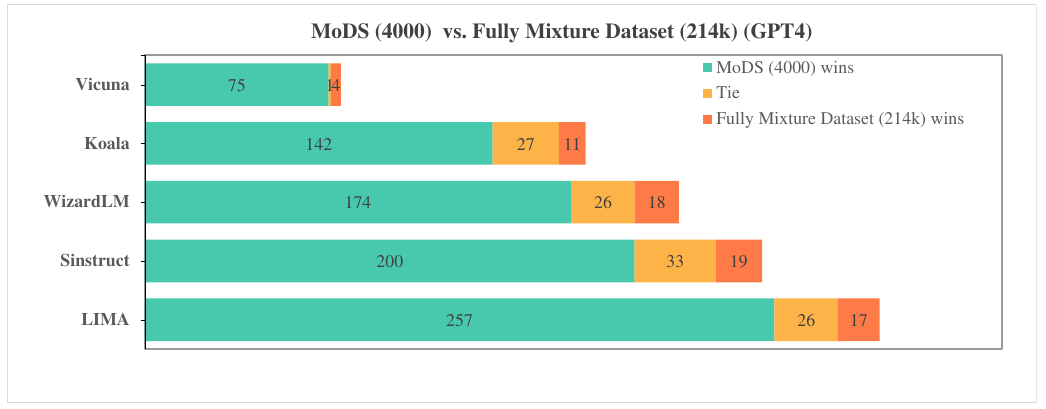}
	\caption{The comparison of our MoDS model trained on selected data with the model trained on full mixture dataset.}
	\label{fig:mods-vs-fullyMix}
\end{figure}

In addition to the Alpaca dataset, Figure \ref{fig:mods-vs-fullyMix}  presents the comparison results of our MoDS model trained on the selected data with the model trained on full Mixture Instruction Dataset. While fine-tuning on this dataset, the size of seed instructions and augmented instructions of our MoDS model are 1000 and 3000 respectively. From this figure, we can see that MoDS performs significantly better than the model trained on full mixture dataset. However, our MoDS model only adopt 4,000 instructions to fine-tuning the pre-trained language model while the model trained on full Mixture Dataset utilize 214K instructions. This result once again demonstrates that our proposed approach could effectively select valuable instruction data from large-scale datasets for a target LLM.

\subsection{Ablation Study}

\begin{figure}[!t]
	\centering
	\includegraphics[width=\linewidth]{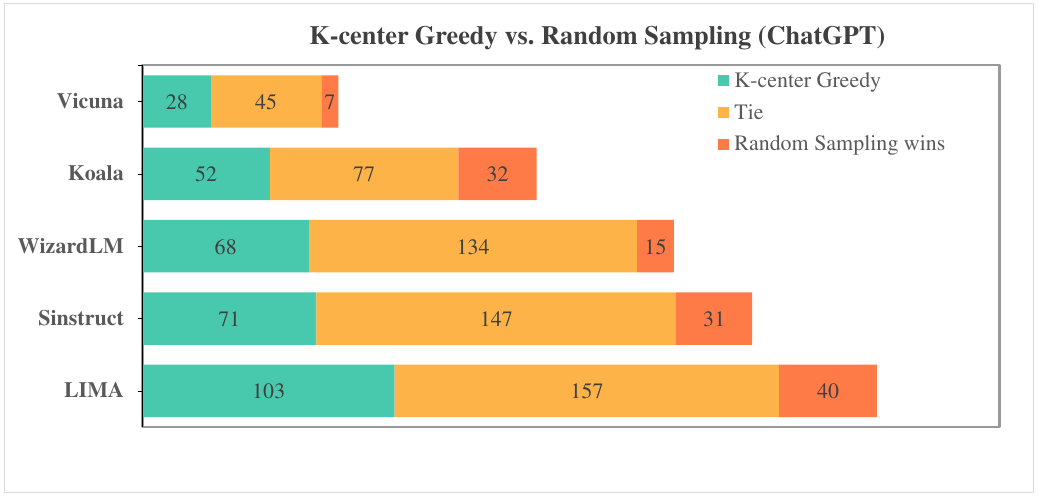}
	\caption{The comparison of models fine-tuned with K-center greedy algorithm and random sampling approach respectively on Mixture Dataset.}
	\label{fig:greedy-random}
\end{figure}

To select instruction data with maximum coverage, this paper proposes to use K-center greedy algorithm to select data from high-quality datasets. In order to analyze the effect of K-center greedy algorithm on data selection, Figure \ref{fig:greedy-random} shows the comparison results of K-center greedy and random sampling approaches on diffferent testsets. In this figure, we first select data from the high-quality dataset with K-center greedy algorithm and random sampling approach respectively, and then fine-tune the pre-trained language model with the selected subsets. The number of selected instruction is 1000, and the original instruction dataset is the Mixture Dataset which includes 214k instrucitons. From this figure, we can see that the model fine-tuned with K-center greedy algorithm performs much better than the model which is fine-tuned with random sampling approach. It indicates that K-center greedy algorithm could select more valuable and diverse instruction data from high-quality dataset.

\begin{figure}[!t]
	\centering
	\includegraphics[width=\linewidth]{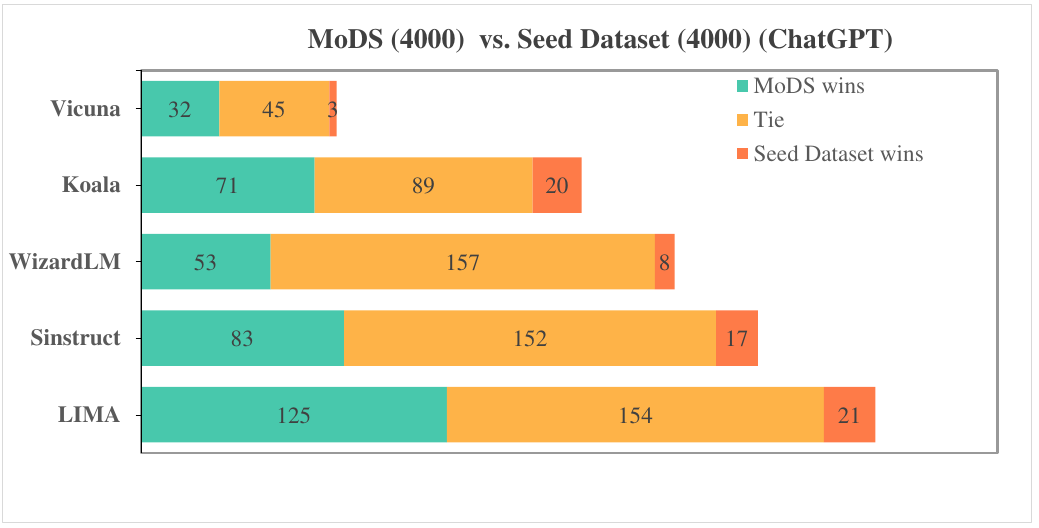}
	\caption{The comparison of MoDS model and the model which is just fine-tuned with seed instruction data of Mixture Dataset.}
	\label{fig:figure4}
\end{figure}

In Figure \ref{fig:figure4}, we compare MoDS with the model which is just fine-tuned with the seed instruction data extracted from Mixture Dataset. Through this way, we can check whether the augmented instruction data could further improve the ability of LLMs. Instead of selecting 1,000 seed instruciton data and 3,000 augmented instruction data respectively, in Figure \ref{fig:figure4} we directly select 4,000 seed instruction data from the high-quality subset of Mixture Dataset. After that, we utilize these 4,000 instruction data to fine-tune the pre-trained language model and compare its performance with MoDS. From this figure, we can see that MoDS is much better than the model fine-tuned with 4,000 seed instruction data. This result demonstrates that the augmented instruction data could effectively compensate for LLM's capacity gaps, thus further enhancing its instruction-following capability.

\begin{figure}[!t]
	\centering
	\includegraphics[width=\linewidth]{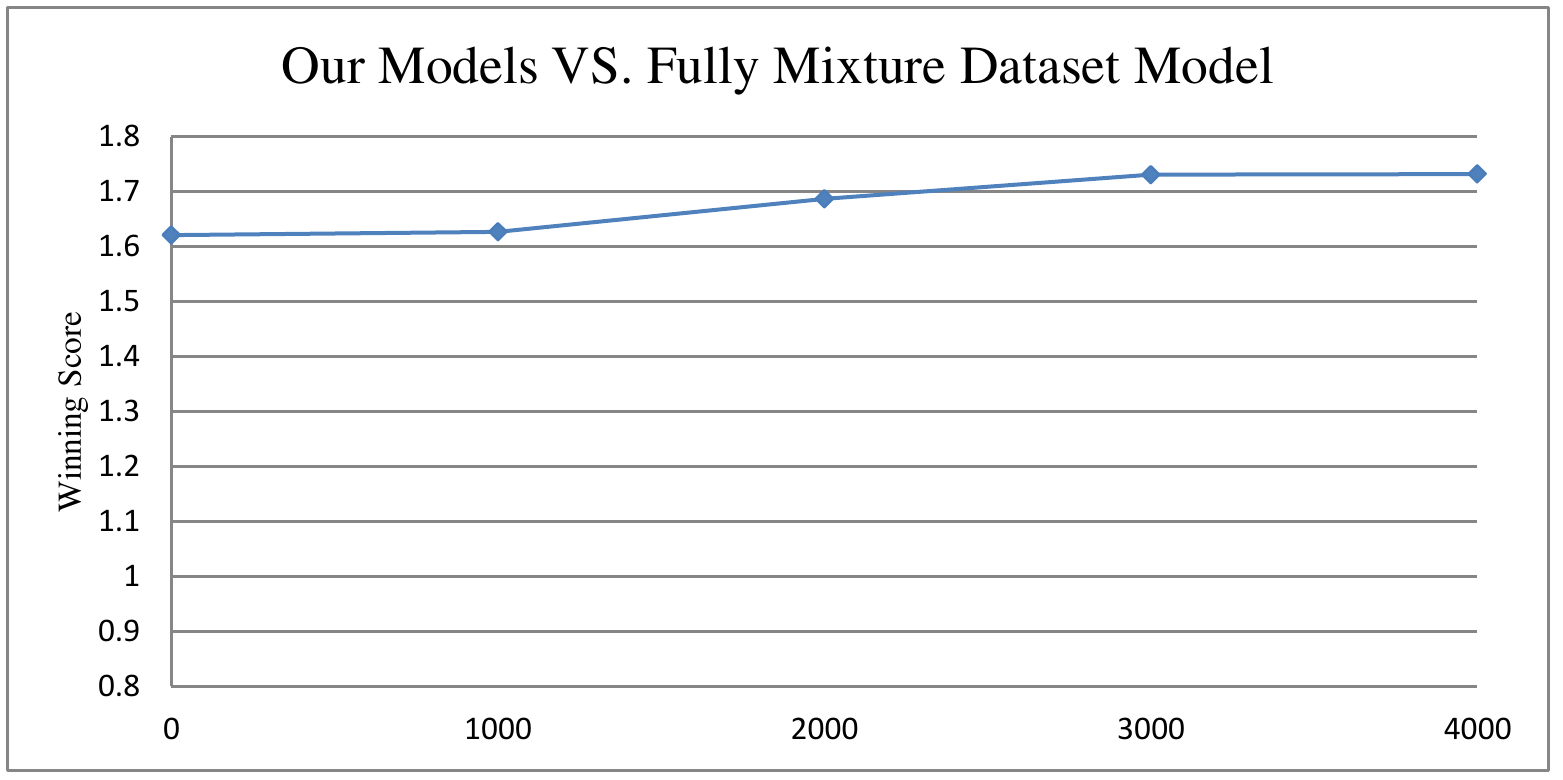}
	\caption{The winning scores of our models with different number of augmented data on Mixture Dataset. All the comparisons of these models are judged by ChatGPT.}
	\label{fig:figure5}
\end{figure}

To investigate the impact of instruction number on LLMs in our approach, Figure \ref{fig:figure5} presents the winning scores of our models with different numbers of augmented instruction data on Mixture Dataset. Following the work of \cite{li2023quantity}, the winning score is also computed by (Num(win)-Num(lose)/Num(all)) + 1. The number of "win", "lose" and "all" are also computed across all five test sets. And the values of winning score which are higher than 1.0 represents our model performs better than the model fine-tuned with full Mixture Dataset, while the values below 1.0 indicate that our model's performance is worse than the full Mixture Dataset model. From this figure, we can see that the performance of our models could effectively improve when we increase the number of augmented instruction data. This result also illustrates that the augmented data are very valuable to enhance the instruction-following capabilities of LLMs.   Furthermore, when the size of augmented dataset reaches 3000, the performance of the model no longer significantly improves. This result suggests that using 3,000 augmented instruction data for Mixture Dataset is already enough in compensating for the model's capability shortcomings.

\section{Conclustion}
\label{sec:bibtex}

In this paper, we propose a model-oriented instruction data selection approach to select valuable instructions for a target foundation LLM. During the selection of instruction data, our approach not only considers the quality and coverage of instruction data, but also integrates the necessity of instructions based on the ability of target LLM. First of all, in our approach, we use a quality evaluation model to evaluate all the (instruction, input, output) triplets in the datasets, and then filter out the instructions with high quality. Secondly, we use a K-center greedy algorithm to select a seed instruction dataset from the high-quality dataset, which makes the selected data as diverse as possible and have a broad coverage. Thirdly, we use the seed instruction dataset to fine-tune the foundation LLM, and then evaluate the fine-tuned LLM on all high-quality instructions to find out the augmented instruction data for the target LLM, which could effectively compensate for the model's capability gaps. Finally, by merging the seed instruction data and the augmented data, we can get a high-quality, broad-coverage and high-necessity dataset from the original large-scale datasets. The final selection dataset is used to fine-tune the foundation LLM to generate the optimized LLM which have the powerful instruction-following capability.

\bibliography{anthology,custom}

\begin{thebibliography}{39}
\expandafter\ifx\csname natexlab\endcsname\relax\def\natexlab#1{#1}\fi

\bibitem[{Bai et~al.(2022)Bai, Jones, Ndousse, Askell, Chen, DasSarma, Drain,
  Fort, Ganguli, Henighan, Joseph, Kadavath, Kernion, Conerly, El-Showk,
  Elhage, Hatfield-Dodds, Hernandez, Hume, Johnston, Kravec, Lovitt, Nanda,
  Olsson, Amodei, Brown, Clark, McCandlish, Olah, Mann, and
  Kaplan}]{bai2022training}
Yuntao Bai, Andy Jones, Kamal Ndousse, Amanda Askell, Anna Chen, Nova DasSarma,
  Dawn Drain, Stanislav Fort, Deep Ganguli, Tom Henighan, Nicholas Joseph,
  Saurav Kadavath, Jackson Kernion, Tom Conerly, Sheer El-Showk, Nelson Elhage,
  Zac Hatfield-Dodds, Danny Hernandez, Tristan Hume, Scott Johnston, Shauna
  Kravec, Liane Lovitt, Neel Nanda, Catherine Olsson, Dario Amodei, Tom Brown,
  Jack Clark, Sam McCandlish, Chris Olah, Ben Mann, and Jared Kaplan. 2022.
\newblock \href {http://arxiv.org/abs/2204.05862} {Training a helpful and
  harmless assistant with reinforcement learning from human feedback}.

\bibitem[{Brown et~al.(2020)Brown, Mann, Ryder, Subbiah, Kaplan, Dhariwal,
  Neelakantan, Shyam, Sastry, Askell et~al.}]{brown2020language}
Tom Brown, Benjamin Mann, Nick Ryder, Melanie Subbiah, Jared~D Kaplan, Prafulla
  Dhariwal, Arvind Neelakantan, Pranav Shyam, Girish Sastry, Amanda Askell,
  et~al. 2020.
\newblock Language models are few-shot learners.
\newblock \emph{Advances in neural information processing systems},
  33:1877--1901.

\bibitem[{Cao et~al.(2023)Cao, Kang, and Sun}]{cao2023instruction}
Yihan Cao, Yanbin Kang, and Lichao Sun. 2023.
\newblock \href {http://arxiv.org/abs/2307.06290} {Instruction mining:
  High-quality instruction data selection for large language models}.

\bibitem[{Chen et~al.(2023)Chen, Li, Yan, Wang, Gunaratna, Yadav, Tang,
  Srinivasan, Zhou, Huang, and Jin}]{chen2023alpagasus}
Lichang Chen, Shiyang Li, Jun Yan, Hai Wang, Kalpa Gunaratna, Vikas Yadav,
  Zheng Tang, Vijay Srinivasan, Tianyi Zhou, Heng Huang, and Hongxia Jin. 2023.
\newblock \href {http://arxiv.org/abs/2307.08701} {Alpagasus: Training a better
  alpaca with fewer data}.

\bibitem[{Chiang et~al.(2023)Chiang, Li, Lin, Sheng, Wu, Zhang, Zheng, Zhuang,
  Zhuang, Gonzalez, Stoica, and Xing}]{vicuna2023}
Wei-Lin Chiang, Zhuohan Li, Zi~Lin, Ying Sheng, Zhanghao Wu, Hao Zhang, Lianmin
  Zheng, Siyuan Zhuang, Yonghao Zhuang, Joseph~E. Gonzalez, Ion Stoica, and
  Eric~P. Xing. 2023.
\newblock \href {https://lmsys.org/blog/2023-03-30-vicuna/} {Vicuna: An
  open-source chatbot impressing gpt-4 with 90\%* chatgpt quality}.

\bibitem[{Chowdhery et~al.(2022)Chowdhery, Narang, Devlin, Bosma, Mishra,
  Roberts, Barham, Chung, Sutton, Gehrmann et~al.}]{chowdhery2022palm}
Aakanksha Chowdhery, Sharan Narang, Jacob Devlin, Maarten Bosma, Gaurav Mishra,
  Adam Roberts, Paul Barham, Hyung~Won Chung, Charles Sutton, Sebastian
  Gehrmann, et~al. 2022.
\newblock Palm: Scaling language modeling with pathways.
\newblock \emph{arXiv preprint arXiv:2204.02311}.

\bibitem[{Conover et~al.(2023)Conover, Hayes, Mathur, Xie, Wan, Shah, Ghodsi,
  Wendell, Zaharia, and Xin}]{DatabricksBlog2023DollyV2}
Mike Conover, Matt Hayes, Ankit Mathur, Jianwei Xie, Jun Wan, Sam Shah, Ali
  Ghodsi, Patrick Wendell, Matei Zaharia, and Reynold Xin. 2023.
\newblock \href
  {https://www.databricks.com/blog/2023/04/12/dolly-first-open-commercially-viable-instruction-tuned-llm}
  {Free dolly: Introducing the world's first truly open instruction-tuned llm}.

\bibitem[{Devlin et~al.(2018)Devlin, Chang, Lee, and
  Toutanova}]{devlin2018bert}
Jacob Devlin, Ming-Wei Chang, Kenton Lee, and Kristina Toutanova. 2018.
\newblock Bert: Pre-training of deep bidirectional transformers for language
  understanding.
\newblock \emph{arXiv preprint arXiv:1810.04805}.

\bibitem[{Ding et~al.(2023)Ding, Chen, Xu, Qin, Zheng, Hu, Liu, Sun, and
  Zhou}]{ding2023enhancing}
Ning Ding, Yulin Chen, Bokai Xu, Yujia Qin, Zhi Zheng, Shengding Hu, Zhiyuan
  Liu, Maosong Sun, and Bowen Zhou. 2023.
\newblock Enhancing chat language models by scaling high-quality instructional
  conversations.
\newblock \emph{arXiv preprint arXiv:2305.14233}.

\bibitem[{Guo et~al.()Guo, Zhang, Wang, Jiang, Nie, Ding, Yue, and
  Wu}]{guo-etal-2023-hc3}
Biyang Guo, Xin Zhang, Ziyuan Wang, Minqi Jiang, Jinran Nie, Yuxuan Ding,
  Jianwei Yue, and Yupeng Wu.
\newblock How close is chatgpt to human experts? comparison corpus, evaluation,
  and detection.
\newblock \emph{arXiv preprint arxiv:2301.07597}.

\bibitem[{He et~al.(2021)He, Liu, Gao, and Chen}]{he2021deberta}
Pengcheng He, Xiaodong Liu, Jianfeng Gao, and Weizhu Chen. 2021.
\newblock \href {http://arxiv.org/abs/2006.03654} {Deberta: Decoding-enhanced
  bert with disentangled attention}.

\bibitem[{Honovich et~al.(2022)Honovich, Scialom, Levy, and
  Schick}]{honovich2022unnatural}
Or~Honovich, Thomas Scialom, Omer Levy, and Timo Schick. 2022.
\newblock Unnatural instructions: Tuning language models with (almost) no human
  labor.
\newblock \emph{arXiv preprint arXiv:2212.09689}.

\bibitem[{Iyer et~al.(2023)Iyer, Lin, Pasunuru, Mihaylov, Simig, Yu, Shuster,
  Wang, Liu, Koura, Li, O'Horo, Pereyra, Wang, Dewan, Celikyilmaz, Zettlemoyer,
  and Stoyanov}]{iyer2023optiml}
Srinivasan Iyer, Xi~Victoria Lin, Ramakanth Pasunuru, Todor Mihaylov, Daniel
  Simig, Ping Yu, Kurt Shuster, Tianlu Wang, Qing Liu, Punit~Singh Koura, Xian
  Li, Brian O'Horo, Gabriel Pereyra, Jeff Wang, Christopher Dewan, Asli
  Celikyilmaz, Luke Zettlemoyer, and Ves Stoyanov. 2023.
\newblock \href {http://arxiv.org/abs/2212.12017} {Opt-iml: Scaling language
  model instruction meta learning through the lens of generalization}.

\bibitem[{K{\"o}pf et~al.(2023)K{\"o}pf, Kilcher, von R{\"u}tte, Anagnostidis,
  Tam, Stevens, Barhoum, Duc, Stanley, Nagyfi et~al.}]{kopf2023openassistant}
Andreas K{\"o}pf, Yannic Kilcher, Dimitri von R{\"u}tte, Sotiris Anagnostidis,
  Zhi-Rui Tam, Keith Stevens, Abdullah Barhoum, Nguyen~Minh Duc, Oliver
  Stanley, Rich{\'a}rd Nagyfi, et~al. 2023.
\newblock Openassistant conversations--democratizing large language model
  alignment.
\newblock \emph{arXiv preprint arXiv:2304.07327}.

\bibitem[{Li et~al.(2023{\natexlab{a}})Li, Zhang, Li, Chen, Chen, Cheng, Wang,
  Zhou, and Xiao}]{li2023quantity}
Ming Li, Yong Zhang, Zhitao Li, Jiuhai Chen, Lichang Chen, Ning Cheng, Jianzong
  Wang, Tianyi Zhou, and Jing Xiao. 2023{\natexlab{a}}.
\newblock From quantity to quality: Boosting llm performance with self-guided
  data selection for instruction tuning.
\newblock \emph{arXiv preprint arXiv:2308.12032}.

\bibitem[{Li et~al.(2023{\natexlab{b}})Li, Yu, Zhou, Schick, Zettlemoyer, Levy,
  Weston, and Lewis}]{li2023self}
Xian Li, Ping Yu, Chunting Zhou, Timo Schick, Luke Zettlemoyer, Omer Levy,
  Jason Weston, and Mike Lewis. 2023{\natexlab{b}}.
\newblock Self-alignment with instruction backtranslation.
\newblock \emph{arXiv preprint arXiv:2308.06259}.

\bibitem[{Longpre et~al.(2023)Longpre, Hou, Vu, Webson, Chung, Tay, Zhou, Le,
  Zoph, Wei, and Roberts}]{longpre2023flan}
Shayne Longpre, Le~Hou, Tu~Vu, Albert Webson, Hyung~Won Chung, Yi~Tay, Denny
  Zhou, Quoc~V. Le, Barret Zoph, Jason Wei, and Adam Roberts. 2023.
\newblock \href {http://arxiv.org/abs/2301.13688} {The flan collection:
  Designing data and methods for effective instruction tuning}.

\bibitem[{Nakano et~al.(2022)Nakano, Hilton, Balaji, Wu, Ouyang, Kim, Hesse,
  Jain, Kosaraju, Saunders, Jiang, Cobbe, Eloundou, Krueger, Button, Knight,
  Chess, and Schulman}]{nakano2022webgpt}
Reiichiro Nakano, Jacob Hilton, Suchir Balaji, Jeff Wu, Long Ouyang, Christina
  Kim, Christopher Hesse, Shantanu Jain, Vineet Kosaraju, William Saunders,
  Xu~Jiang, Karl Cobbe, Tyna Eloundou, Gretchen Krueger, Kevin Button, Matthew
  Knight, Benjamin Chess, and John Schulman. 2022.
\newblock \href {http://arxiv.org/abs/2112.09332} {Webgpt: Browser-assisted
  question-answering with human feedback}.

\bibitem[{Ni et~al.(2023)Ni, Xue, Jain, Shah, Zheng, and You}]{instructionwild}
Jinjie Ni, Fuzhao Xue, Kabir Jain, Mahir~Hitesh Shah, Zangwei Zheng, and Yang
  You. 2023.
\newblock Instruction in the wild: A user-based instruction dataset.
\newblock \url{https://github.com/XueFuzhao/InstructionWild}.

\bibitem[{OpenAI(2023)}]{GPT4}
OpenAI. 2023.
\newblock {GPT-4} technical report.
\newblock \emph{CoRR}, abs/2303.08774.

\bibitem[{Ouyang et~al.(2022)Ouyang, Wu, Jiang, Almeida, Wainwright, Mishkin,
  Zhang, Agarwal, Slama, Ray et~al.}]{ouyang2022training}
Long Ouyang, Jeffrey Wu, Xu~Jiang, Diogo Almeida, Carroll Wainwright, Pamela
  Mishkin, Chong Zhang, Sandhini Agarwal, Katarina Slama, Alex Ray, et~al.
  2022.
\newblock Training language models to follow instructions with human feedback.
\newblock \emph{Advances in Neural Information Processing Systems},
  35:27730--27744.

\bibitem[{Sener and Savarese(2017)}]{sener2017active}
Ozan Sener and Silvio Savarese. 2017.
\newblock Active learning for convolutional neural networks: A core-set
  approach.
\newblock \emph{arXiv preprint arXiv:1708.00489}.

\bibitem[{Shu et~al.(2023)Shu, Wang, Zhu, Geiping, Xiao, and
  Goldstein}]{shu2023exploitability}
Manli Shu, Jiongxiao Wang, Chen Zhu, Jonas Geiping, Chaowei Xiao, and Tom
  Goldstein. 2023.
\newblock On the exploitability of instruction tuning.
\newblock \emph{arXiv preprint arXiv:2306.17194}.

\bibitem[{Stiennon et~al.(2022)Stiennon, Ouyang, Wu, Ziegler, Lowe, Voss,
  Radford, Amodei, and Christiano}]{stiennon2022learning}
Nisan Stiennon, Long Ouyang, Jeff Wu, Daniel~M. Ziegler, Ryan Lowe, Chelsea
  Voss, Alec Radford, Dario Amodei, and Paul Christiano. 2022.
\newblock \href {http://arxiv.org/abs/2009.01325} {Learning to summarize from
  human feedback}.

\bibitem[{Sun et~al.(2023)Sun, Shen, Zhou, Zhang, Chen, Cox, Yang, and
  Gan}]{sun2023principle}
Zhiqing Sun, Yikang Shen, Qinhong Zhou, Hongxin Zhang, Zhenfang Chen, David
  Cox, Yiming Yang, and Chuang Gan. 2023.
\newblock Principle-driven self-alignment of language models from scratch with
  minimal human supervision.
\newblock \emph{arXiv preprint arXiv:2305.03047}.

\bibitem[{Taori et~al.(2023)Taori, Gulrajani, Zhang, Dubois, Li, Guestrin,
  Liang, and Hashimoto}]{alpaca}
Rohan Taori, Ishaan Gulrajani, Tianyi Zhang, Yann Dubois, Xuechen Li, Carlos
  Guestrin, Percy Liang, and Tatsunori~B. Hashimoto. 2023.
\newblock Stanford alpaca: An instruction-following llama model.
\newblock \url{https://github.com/tatsu-lab/stanford_alpaca}.

\bibitem[{Touvron et~al.(2023{\natexlab{a}})Touvron, Lavril, Izacard, Martinet,
  Lachaux, Lacroix, Rozi{\`{e}}re, Goyal, Hambro, Azhar, Rodriguez, Joulin,
  Grave, and Lample}]{llama}
Hugo Touvron, Thibaut Lavril, Gautier Izacard, Xavier Martinet, Marie{-}Anne
  Lachaux, Timoth{\'{e}}e Lacroix, Baptiste Rozi{\`{e}}re, Naman Goyal, Eric
  Hambro, Faisal Azhar, Aur{\'{e}}lien Rodriguez, Armand Joulin, Edouard Grave,
  and Guillaume Lample. 2023{\natexlab{a}}.
\newblock Llama: Open and efficient foundation language models.
\newblock \emph{CoRR}, abs/2302.13971.

\bibitem[{Touvron et~al.(2023{\natexlab{b}})Touvron, Martin, Stone, Albert,
  Almahairi, Babaei, Bashlykov, Batra, Bhargava, Bhosale
  et~al.}]{touvron2023llama}
Hugo Touvron, Louis Martin, Kevin Stone, Peter Albert, Amjad Almahairi, Yasmine
  Babaei, Nikolay Bashlykov, Soumya Batra, Prajjwal Bhargava, Shruti Bhosale,
  et~al. 2023{\natexlab{b}}.
\newblock Llama 2: Open foundation and fine-tuned chat models.
\newblock \emph{arXiv preprint arXiv:2307.09288}.

\bibitem[{Vu et~al.(2023)Vu, He, Haffari, and Shareghi}]{vu2023koala}
Thuy-Trang Vu, Xuanli He, Gholamreza Haffari, and Ehsan Shareghi. 2023.
\newblock Koala: An index for quantifying overlaps with pre-training corpora.
\newblock \emph{arXiv preprint arXiv:2303.14770}.

\bibitem[{Wang et~al.(2023{\natexlab{a}})Wang, Cheng, Zhan, Li, Song, and
  Liu}]{wang2023openchat}
Guan Wang, Sijie Cheng, Xianyuan Zhan, Xiangang Li, Sen Song, and Yang Liu.
  2023{\natexlab{a}}.
\newblock Openchat: Advancing open-source language models with mixed-quality
  data.
\newblock \emph{arXiv preprint arXiv:2309.11235}.

\bibitem[{Wang et~al.(2023{\natexlab{b}})Wang, Ivison, Dasigi, Hessel, Khot,
  Chandu, Wadden, MacMillan, Smith, Beltagy, and Hajishirzi}]{wang2023far}
Yizhong Wang, Hamish Ivison, Pradeep Dasigi, Jack Hessel, Tushar Khot,
  Khyathi~Raghavi Chandu, David Wadden, Kelsey MacMillan, Noah~A. Smith,
  Iz~Beltagy, and Hannaneh Hajishirzi. 2023{\natexlab{b}}.
\newblock \href {http://arxiv.org/abs/2306.04751} {How far can camels go?
  exploring the state of instruction tuning on open resources}.

\bibitem[{Wang et~al.(2022{\natexlab{a}})Wang, Kordi, Mishra, Liu, Smith,
  Khashabi, and Hajishirzi}]{wang2022self}
Yizhong Wang, Yeganeh Kordi, Swaroop Mishra, Alisa Liu, Noah~A Smith, Daniel
  Khashabi, and Hannaneh Hajishirzi. 2022{\natexlab{a}}.
\newblock Self-instruct: Aligning language model with self generated
  instructions.
\newblock \emph{arXiv preprint arXiv:2212.10560}.

\bibitem[{Wang et~al.(2022{\natexlab{b}})Wang, Mishra, Alipoormolabashi, Kordi,
  Mirzaei, Arunkumar, Ashok, Dhanasekaran, Naik, Stap, Pathak, Karamanolakis,
  Lai, Purohit, Mondal, Anderson, Kuznia, Doshi, Patel, Pal, Moradshahi,
  Parmar, Purohit, Varshney, Kaza, Verma, Puri, Karia, Sampat, Doshi, Mishra,
  Reddy, Patro, Dixit, Shen, Baral, Choi, Smith, Hajishirzi, and
  Khashabi}]{wang2022supernaturalinstructions}
Yizhong Wang, Swaroop Mishra, Pegah Alipoormolabashi, Yeganeh Kordi, Amirreza
  Mirzaei, Anjana Arunkumar, Arjun Ashok, Arut~Selvan Dhanasekaran, Atharva
  Naik, David Stap, Eshaan Pathak, Giannis Karamanolakis, Haizhi~Gary Lai,
  Ishan Purohit, Ishani Mondal, Jacob Anderson, Kirby Kuznia, Krima Doshi,
  Maitreya Patel, Kuntal~Kumar Pal, Mehrad Moradshahi, Mihir Parmar, Mirali
  Purohit, Neeraj Varshney, Phani~Rohitha Kaza, Pulkit Verma, Ravsehaj~Singh
  Puri, Rushang Karia, Shailaja~Keyur Sampat, Savan Doshi, Siddhartha Mishra,
  Sujan Reddy, Sumanta Patro, Tanay Dixit, Xudong Shen, Chitta Baral, Yejin
  Choi, Noah~A. Smith, Hannaneh Hajishirzi, and Daniel Khashabi.
  2022{\natexlab{b}}.
\newblock \href {http://arxiv.org/abs/2204.07705} {Super-naturalinstructions:
  Generalization via declarative instructions on 1600+ nlp tasks}.

\bibitem[{Wang et~al.(2023{\natexlab{c}})Wang, Zhong, Li, Mi, Zeng, Huang,
  Shang, Jiang, and Liu}]{wang2023aligning}
Yufei Wang, Wanjun Zhong, Liangyou Li, Fei Mi, Xingshan Zeng, Wenyong Huang,
  Lifeng Shang, Xin Jiang, and Qun Liu. 2023{\natexlab{c}}.
\newblock Aligning large language models with human: A survey.
\newblock \emph{arXiv preprint arXiv:2307.12966}.

\bibitem[{Wei et~al.(2021)Wei, Bosma, Zhao, Guu, Yu, Lester, Du, Dai, and
  Le}]{wei2021finetuned}
Jason Wei, Maarten Bosma, Vincent~Y Zhao, Kelvin Guu, Adams~Wei Yu, Brian
  Lester, Nan Du, Andrew~M Dai, and Quoc~V Le. 2021.
\newblock Finetuned language models are zero-shot learners.
\newblock \emph{arXiv preprint arXiv:2109.01652}.

\bibitem[{Xu et~al.(2023)Xu, Sun, Zheng, Geng, Zhao, Feng, Tao, and
  Jiang}]{xu2023wizardlm}
Can Xu, Qingfeng Sun, Kai Zheng, Xiubo Geng, Pu~Zhao, Jiazhan Feng, Chongyang
  Tao, and Daxin Jiang. 2023.
\newblock Wizardlm: Empowering large language models to follow complex
  instructions.
\newblock \emph{arXiv preprint arXiv:2304.12244}.

\bibitem[{Yu et~al.(2023)Yu, Zhuang, Zhang, Meng, Ratner, Krishna, Shen, and
  Zhang}]{yu2023large}
Yue Yu, Yuchen Zhuang, Jieyu Zhang, Yu~Meng, Alexander Ratner, Ranjay Krishna,
  Jiaming Shen, and Chao Zhang. 2023.
\newblock Large language model as attributed training data generator: A tale of
  diversity and bias.
\newblock \emph{arXiv preprint arXiv:2306.15895}.

\bibitem[{Zhang et~al.(2022)Zhang, Roller, Goyal, Artetxe, Chen, Chen, Dewan,
  Diab, Li, Lin et~al.}]{zhang2022opt}
Susan Zhang, Stephen Roller, Naman Goyal, Mikel Artetxe, Moya Chen, Shuohui
  Chen, Christopher Dewan, Mona Diab, Xian Li, Xi~Victoria Lin, et~al. 2022.
\newblock Opt: Open pre-trained transformer language models.
\newblock \emph{arXiv preprint arXiv:2205.01068}.

\bibitem[{Zhou et~al.(2023)Zhou, Liu, Xu, Iyer, Sun, Mao, Ma, Efrat, Yu, Yu
  et~al.}]{zhou2023lima}
Chunting Zhou, Pengfei Liu, Puxin Xu, Srini Iyer, Jiao Sun, Yuning Mao, Xuezhe
  Ma, Avia Efrat, Ping Yu, Lili Yu, et~al. 2023.
\newblock Lima: Less is more for alignment.
\newblock \emph{arXiv preprint arXiv:2305.11206}.

\end{thebibliography}
\bibliographystyle{acl_natbib}

\appendix



\end{document}